\documentclass[11pt,a4paper]{article}
\usepackage[hyperref]{acl2021}
\usepackage{times}
\usepackage{latexsym}
\usepackage{subcaption, booktabs}  
\usepackage{graphicx}
\usepackage{amsmath}
\usepackage{bm}
\usepackage{xcolor}
\usepackage{soul}
\usepackage{tikz}

\newcommand{\inline}[2]{%
    \begin{tikzpicture}[baseline=(word.base), txt/.style={shape=rectangle, inner sep=0pt}]
    \node[txt] (word) {#1};
    \node[above] at (word.north) {\footnotesize{#2}};
    \end{tikzpicture}%
    }

\newcommand{\hlc}[2][yellow]{{%
    \colorlet{foo}{#1}%
    \sethlcolor{foo}\hl{#2}}%
}

\usepackage{microtype}

\aclfinalcopy 
\def\aclpaperid{2807} 

\newcommand{\xv}{\ensuremath{\bm{x}}}
\newcommand{\yv}{\ensuremath{\bm{y}}}

\title{Do Context-Aware Translation Models Pay the Right Attention?}

\author{
Kayo Yin$^{1}$  \quad
Patrick Fernandes$^{1,2,3}$  \quad
Danish Pruthi$^{1}$  \quad
Aditi Chaudhary$^{1}$  \\
\textbf{André F. T. Martins}$^{2,3,4}$ \quad
\textbf{Graham Neubig}$^{1}$ \\
$^1$Language Technologies Institute, Carnegie Mellon University, Pittsburgh, PA \\
$^2$Instituto Superior Técnico \& LUMLIS (Lisbon ELLIS Unit), Lisbon, Portugal \\
$^3$Instituto de Telecomunicações, Lisbon, Portugal \\
$^4$Unbabel, Lisbon, Portugal \\
 \texttt{\{kayoy, pfernand, ddanish, aschaudh, gneubig\}@cs.cmu.edu}  \\ \texttt{andre.t.martins@tecnico.ulisboa.pt}
}

\date{}

\begin{document}
\maketitle
\begin{abstract}
Context-aware machine translation models are designed to leverage contextual information, but often fail to do so. As a result, they inaccurately disambiguate pronouns and polysemous words that require context for resolution.
In this paper, we ask several questions: What contexts do human translators use to resolve ambiguous words? Are models paying large amounts of attention to the same context? What if we explicitly train them to do so? 
To answer these questions, we introduce \textbf{\texttt{SCAT}} (Supporting Context for Ambiguous Translations), a new English-French 
dataset comprising supporting context words for 14K  translations that professional translators found useful for pronoun disambiguation.
Using \textbf{\texttt{SCAT}}, we perform an in-depth analysis of the context used to disambiguate, examining positional and lexical characteristics of the supporting words.
Furthermore, we measure the degree of alignment between the model's attention scores and the supporting context from \textbf{\texttt{SCAT}}, and apply a guided attention strategy to encourage agreement between the two.%
\footnote{Our \textbf{\texttt{SCAT}} data and code for experiments are available at \url{https://github.com/neulab/contextual-mt}.}
\end{abstract}

\section{Introduction}


\begin{table}[t!]
\small
\centering
\vspace{-1mm}
\begin{tabular}{@{}cl@{}} \toprule
& \textbf{Human} \\ \midrule
En & \begin{tabular}[l]{@{}l@{}}Look after her a lot. Okay. Any questions?  Have we \\ got her report?  Yes, \textbf{\underline{it}}'s in the infirmary already \end{tabular} \\ 
Fr & \begin{tabular}[l]{@{}l@{}} Dorlotez-la. D'accord. Vous avez des questions ? On \\ dispose de son \hl{rapport} ? Oui, \textbf{\underline{il}} est à l'infirmerie.\\ \end{tabular} \\ \midrule
& \textbf{Context-aware baseline} \\ \midrule
En & \begin{tabular}[l]{@{}l@{}}Look after her a lot. Okay. Any questions?  Have we \\ got her report?  Yes\hlc[cyan!50]{,} \underline{\hlc[cyan!90]{\textbf{it}}}'s in the \hlc[cyan!30]{infirmary} already. \end{tabular} \\ 
Fr & \begin{tabular}[l]{@{}l@{}} Dorlotez-la. D'accord. Vous avez des questions ? On \\ dispose de son rapport \hlc[cyan!90]{?}  \hlc[cyan!50]{Oui,} \hlc[cyan!30]{\textbf{{\textcolor{red}{elle}}}} {est} déjà à l'infirmerie. \\ \end{tabular} \\ \midrule
& \textbf{Model w/ attention regularization} \\ \midrule 
En & \begin{tabular}[l]{@{}l@{}}Look after her a lot. \hlc[cyan!50]{Okay}. Any \hlc[cyan!30]{questions}?  Have we \\ got her \hlc[cyan!90]{report}?  Yes \textbf{\underline{it}}'s in the infirmary already. \end{tabular} \\ 
Fr & \begin{tabular}[l]{@{}l@{}} Dorlotez-la. D'accord. Vous avez des questions ? On \\ dispose de son \hlc[cyan!30]{rapport} ?  \hlc[cyan!50]{Oui}, \hlc[cyan!90]{\textbf{il}} est déjà à l'hôpital \\ \end{tabular} \\ \bottomrule

\end{tabular}
\vspace{-1mm}
\caption{Translation of the ambiguous pronoun ``\underline{it}''. In French, if the referent of ``it'' is masculine (e.g., report) then ``il'' is used, otherwise ``elle''. The model with regularized attention translates the pronoun correctly, with the largest attention on the referent ``report''. Top $3$ words with the highest attention are \hlc[cyan!90]{hig}\hlc[cyan!50]{hlig}\hlc[cyan!30]{hted}.}
\vspace{-4mm}
\label{table:main}
\end{table}

There is a growing consensus in machine translation research that it is necessary to move beyond sentence-level translation and incorporate document-level context \cite{guillou-etal-2018-pronoun, laubli-etal-2018-machine, toral-etal-2018-attaining}.
While various methods to incorporate context in neural machine translation (NMT) have been proposed (\citet{tiedemann-scherrer-2017-neural, miculicich-etal-2018-document, maruf-haffari-2018-document}, \textit{inter alia}), it is unclear whether models rely on the ``right'' context that is actually sufficient to disambiguate difficult translations.
Even when additional context is provided, models often perform poorly on evaluation of relatively simple discourse phenomena \cite{muller-etal-2018-large, bawden-etal-2018-evaluating, voita-etal-2019-good, voita-etal-2019-context, lopes-etal-2020-document} and rely on spurious word co-occurences during translation of polysemous words \cite{emelin2020detecting}. Some evidence suggests that models attend to uninformative tokens \cite{voita-etal-2018-context} and do not use contextual information adequately \cite{kim-etal-2019-document}.

To understand plausibly \textit{why} current NMT models are unable to fully leverage the disambiguating context they are provided, and \textit{how} we can develop models that use context more effectively, we pose the following research questions: (i) In context aware translation, what context is intrinsically useful to disambiguate hard translation phenomena such as ambiguous pronouns or word senses?; (ii) Are context-aware MT models paying attention to the relevant context or not?; and (iii) If not, can we encourage them to do so? 

To answer the first question, we \emph{collect annotations of context that human translators found useful} in choosing between ambiguous translation options (\S\ref{section:human_study}).
Specifically, we ask 20 professional translators to choose the correct French translation between two contrastive translations of an ambiguous word, given an English source sentence and the previous source- and target-side sentences. The translators additionally highlight the words they found the most useful to make their decision, giving an idea of the context useful in making these decisions. We collect 14K such annotations and release \textbf{\texttt{SCAT}} (``Supporting Context for Ambiguous Translations''), the first dataset of human rationales for resolving ambiguity in document-level translation.
Analysis reveals that inter-sentential target context is important for pronoun translation, whereas intra-sentential source context 
is often sufficient for word sense disambiguation.

To answer the second question, we \emph{quantify the similarity of the attention distribution of context-aware models and the human annotations} in \textbf{\texttt{SCAT}} (\S\ref{section:model_attention}). We measure alignment between the baseline context-aware model's attention and human rationales across various model attention heads and layers.
We observe a relatively high alignment between self attention scores from the top encoder layers and the source-side supporting context marked by translators, however, 
the model's attention is poorly aligned with target-side supporting context.  

For the third question, 
we explore a method to \emph{regularize attention towards human-annotated disambiguating context} (\S\ref{section:attn-reg}).
We find that attention regularization is an effective technique to encourage models to pay more attention to words humans find useful to resolve ambiguity in translations. 
Our models with regularized attention outperform previous context-aware baselines, improving translation quality by $0.54$ BLEU, and  yielding a relative improvement of $14.7$\% in contrastive evaluation. An example of translations from a baseline and our model, along with the supporting rationale by a professional translator is illustrated in Table~\ref{table:main}.

\section{Document-Level Translation}
\label{sec:background}

\paragraph{Neural Machine Translation.}
Current NMT models  employ encoder-decoder architectures \cite{bahdanau2014neural, transformer}. First, the encoder maps a {source sequence} $\bm{x} = (x_1, x_2, ..., x_S)$ to a continuous representation $\bm{z} = (z_1, z_2, ..., z_S)$. Then, given $\bm{z}$, the decoder generates the corresponding {target sequence} $\bm{y} = (y_1, y_2, ..., y_T)$, one token at a time. Sentence-level NMT models take one source sentence and generate one target sentence at a time.
These models perform reasonably well, but given that they only have \textbf{intra-sentential context}, they fail to handle some phenomena that require \textbf{inter-sentential context} to accurately translate.
Well-known examples of these phenomena include gender-marked anaphoric pronouns \cite{guillou-etal-2018-pronoun} and maintenance of lexical coherence \cite{laubli-etal-2018-machine}.

\paragraph{Document-Level Translation.}
Document-level translation models learn to maximize the probability of a target document $\bm{Y}$ given the source document $\bm{X}$: $P_\theta (\bm{Y}|\bm{X}) = \prod_{j=1}^JP_\theta(\bm{y}^j | \bm{x}^j, \bm{C}^j)$, where $\bm{y}^j$ and $\bm{x}^j$ are the $j$-th target and source sentences, and $\bm{C}^j$ is the collection of contextual sentences for the $j$-th sentence pair. 
There are many methods for incorporating context (\S\ref{section:related-work}), but even simple \textbf{concatenation} \citep{tiedemann-scherrer-2017-neural}, which prepends the previous source or target sentences to the current sentence separated by a $\langle$\textsc{brk}$\rangle$ tag, achieves comparable performance to more sophisticated approaches, especially in high-resource scenarios \cite{lopes-etal-2020-document}. 

\paragraph{Evaluation.}
BLEU \cite{papineni-etal-2002-bleu} is most widely used to evaluate MT, but it can be poorly correlated with human evaluation \cite{callison-burch-etal-2006-evaluating, reiter-2018-structured}. Recently, a number of neural evaluation methods, such as COMET \cite{rei-etal-2020-comet}, have shown better correlation with human judgement.
Nevertheless, common automatic metrics have limited ability to evaluate discourse in MT \cite{hardmeier2012discourse}. As a remedy to this, researchers often use \textbf{contrastive test sets} for a targeted discourse phenomenon \cite{muller-etal-2018-large}, such as pronoun anaphora resolution and word sense disambiguation, to verify if the model ranks the correct translation of an ambiguous sentence higher than the incorrect translation. 

\section{What Context Do Human Translators Pay Attention to?}
\label{section:human_study}

We first conduct a user study to collect supporting context that translators use in disambiguation, and analyze characteristics of the supporting words.

\subsection{Recruitment and Annotation Setup}
We recruited $20$ freelance English-French translators on Upwork.\footnote{\url{https://www.upwork.com}} The translators %
 are native speakers of at least one of the two languages and have a job success rate of over $90$\%.
Each translator is given $400$ examples with an English source sentence and two possible French translations, and one out of $5$ possible context levels: no context (\textbf{0+0}), only the previous source sentence as context (\textbf{1+0}), only the previous target sentence (\textbf{0+1}), the previous source sentence and target sentence (\textbf{1+1}), and the $5$ previous source and target sentences (\textbf{5+5}). We vary the context level in each example to measure how human translation quality changes. 

\begin{figure}[t]
    \includegraphics[width=\linewidth]{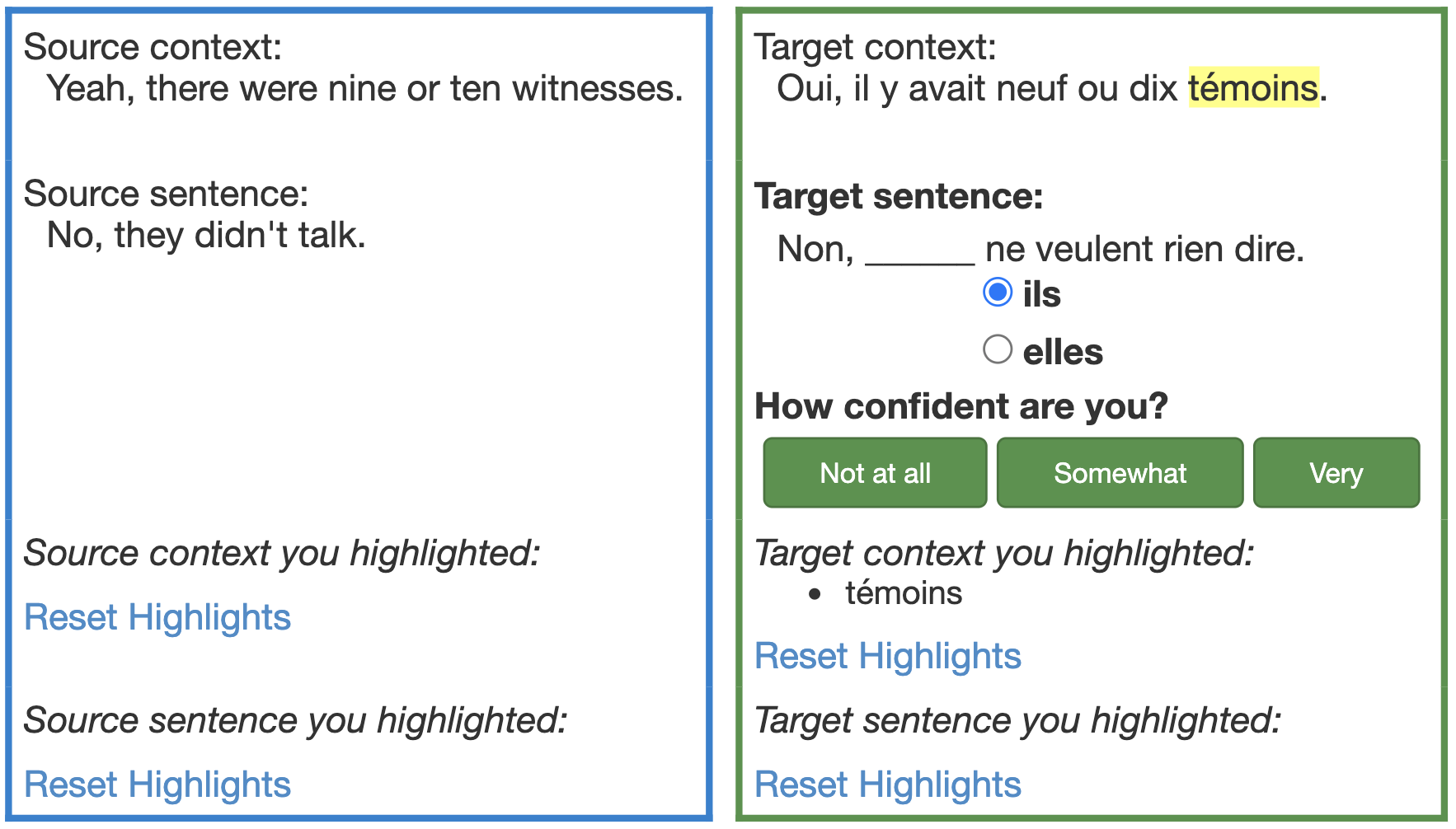}
    \caption{\vspace{-1mm}The annotation page shown to translators.}
    \vspace{-2mm}
    \label{fig:interface}
\end{figure}

Translators provide annotations using the interface shown in Figure \ref{fig:interface}. They are first asked to select the correct translation out of the two contrastive translations, and then highlight word(s) they found useful to arrive at their answer. In cases where multiple words are sufficient to disambiguate, translators were asked to mark only the most salient words rather than all of them.
Further, translators also reported their confidence in their answers, choosing from ``not at all'', ``somewhat'', and ``very''.

\subsection{Tasks and Data Quality}


We perform this study for two tasks: pronoun anaphora resolution (PAR), where the translators are tasked with choosing the correct French gendered pronoun associated to a neutral English pronoun, and word sense disambiguation (WSD), where the translators pick the correct translation of a polysemous word. PAR, and WSD to a lesser extent, have been  commonly studied to evaluate context-aware NMT models  \cite{voita-etal-2018-context,lopes-etal-2020-document, muller-etal-2018-large, huo-etal-2020-diving, nagata-morishita-2020-test}.

\paragraph{Pronoun Anaphora Resolution.}
We annotate examples from the contrastive test set by \newcite{lopes-etal-2020-document}. This set includes $14$K examples from the OpenSubtitles2018 dataset \cite{lison-etal-2018-opensubtitles2018} with occurrences of the English pronouns ``it'' and ``they'' that correspond to the French translations ``il'' or ``elle'' and ``ils'' or ``elles'', with $3.5$K examples for each French pronoun type. 
Through our annotation effort, we obtain 14K examples of supporting context for pronoun anaphora resolution in ambiguous translations selected by professional human translators. Statistics on this dataset, \textbf{\texttt{SCAT}}: \textit{Supporting Context for Ambiguous Translations}, are provided in Appendix \ref{appendix:scat}. 

\paragraph{Word Sense Disambiguation.}
There are no existing contrastive datasets for WSD with a context window larger than $1$ sentence, therefore, we automatically generate contrastive examples with context window of $5$ sentences from OpenSubtitles2018 by identifying polysemous English words and possible French translations. We describe our methodology in Appendix \ref{appendix:wsd}. 


\paragraph{Quality.}
For quality control, we asked $8$ internal speakers of English and French, with native or bilingual proficiency in both languages, to carefully annotate the same 100 examples given to all professional translators. We compared both the answer accuracies and the selected words for each hired translator against this control set and discarded submissions that either had several incorrect answers while the internal bilinguals were able to choose the correct answer on the same example, or that highlighted contextual words that the internal annotators did not select and that had little relevance to the ambiguous word.  
Furthermore, among the $400$ examples given to each annotator, the first hundred are identical, allowing us to measure the inter-annotator agreement for both answer and supporting context selection. 

First, for answer selection on PAR, we find 91.0\% overall agreement, with Fleiss' free-marginal Kappa $\kappa = 0.82$. For WSD, we find 85.9\% overall agreement with $\kappa = 0.72$. This indicates a substantial inter-annotator agreement for the selected answer. 
In addition, we measure the inter-annotator agreement for the selected words by calculating the F1 between the word selections for each pair of annotators given identical context settings. For PAR, we obtain an average F1 of 0.52 across all possible pairs, 
and a standard deviation of 0.12. For WSD, we find an average F1 of 0.46
and a standard deviation of 0.12. There is a high agreement between annotators for the selected words as well.

\begin{table}[t]
\resizebox{\linewidth}{!}{ 
\begin{tabular}{c|cc|cc}
\toprule
& \multicolumn{2}{c|}{PAR} &   \multicolumn{2}{c}{WSD} \\
\toprule
Context & Correct & Not confident & Correct & Not confident \\
\midrule
0 + 0 & 78.4 & 27.0 & 88.7 & 7.0\\
1 + 0 & \bfseries 90.6 & \bfseries 13.2 & 88.7 & 6.5 \\
0 + 1 & \bfseries  93.0 & \bfseries 9.2 & 87.5 & 6.7 \\
1 + 1 & 93.6 & \bfseries 6.7 & 87.1 & 6.5 \\
5 + 5 & \bfseries 95.9 & \bfseries 2.8 & 88.7 & 5.9 \\
\midrule
No ante & 75.4 & 33.8 & \textendash & \textendash \\
Has ante & \bfseries 96.0 & \bfseries 3.3 & \textendash & \textendash\\
\bottomrule
\end{tabular}}
\caption{Percentage of correct and zero-confidence answers by varying context level. $n$+$m$: $n$ previous source and $m$ previous target sentences given as context. Values in bold are significantly different from values above it (p  $<0.05$).}
\label{table:accuracy}
\end{table}

\subsection{Answer Accuracy and Confidence}

Table \ref{table:accuracy} shows the accuracy of answers and the percentage of answers being reported as \textit{not at all} confident for each of the 5 different context levels. For PAR, there is a large increase in accuracy and confidence when just one previous sentence in either language is provided as context compared to no context at all. Target-side context also seems more useful than source: only target-side context gives higher answer accuracy than only source-side context, while the accuracy does not increase significantly by having both previous sentences. 

\begin{figure}
    \centering
    \includegraphics[width=\linewidth]{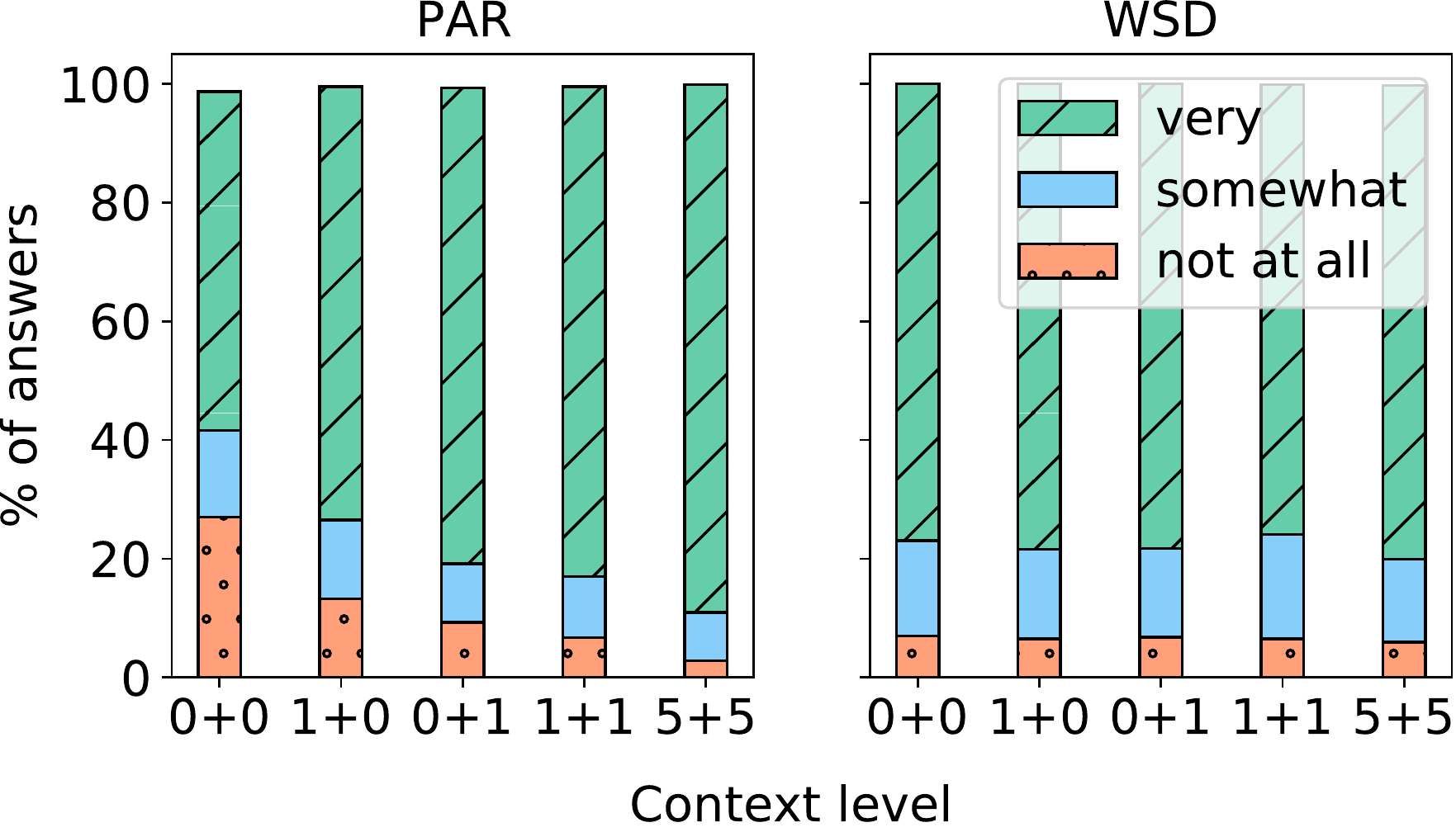}
    \caption{Distribution of confidence in answers per context level for PAR (left) and WSD (right).}
    \label{fig:conf}
\end{figure}

\begin{figure}
    \centering
    \includegraphics[width=\linewidth]{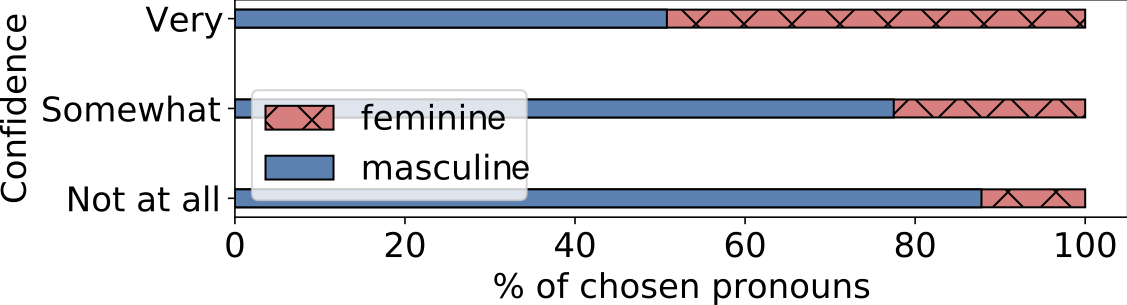}
    \caption{Distribution of gender of selected pronouns per confidence level for PAR.}
    \label{fig:gender}
\end{figure}

For WSD, we do not observe significant differences in answer accuracy and confidence between the different context levels (Figure \ref{fig:conf}).The high answer accuracy with 0+0 context and the low rate of zero-confidence answers across all settings suggest that the necessary disambiguating information is often present in the intra-sentential context. Alternatively, this may be partially due to characteristics of the automatically generated dataset itself: we found that some examples are misaligned so the previous sentences given as context do not actually correspond to the context of the current sentences, and therefore do not add useful information.
We also observe that translators tend to report a high confidence and high agreement in incorrect answers as well. This can be explained by the tendency to select the masculine pronoun in PAR (Figure \ref{fig:gender}) or the prevailing word sense in WSD.  

To properly translate an anaphoric pronoun, the translator must identify its antecedent and determine its gender, so we hypothesize that the antecedent is of high importance for disambiguation.
In our study, $72.4\%$ of the examples shown to annotators contain the antecedent in the context or current sentences. We calculate how answer accuracy and confidence vary between examples that do or do not contain the pronoun antecedent. We find that the presence of the antecedent in the context leads to larger variations in answer accuracy than the level of context given, demonstrating the importance of antecedents for resolution.

\begin{figure}[t]
    \includegraphics[width=\linewidth]{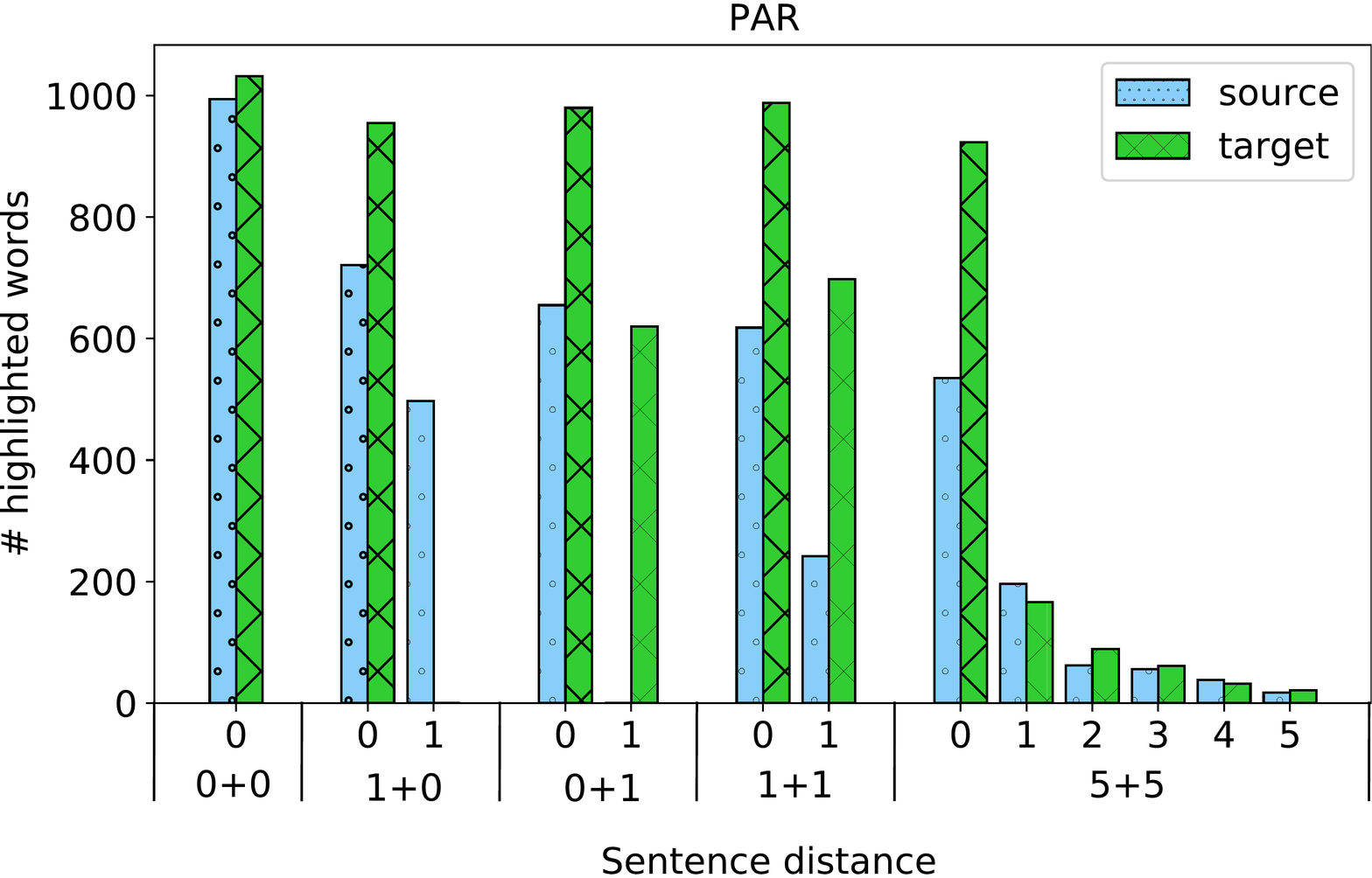}
    \includegraphics[width=\linewidth]{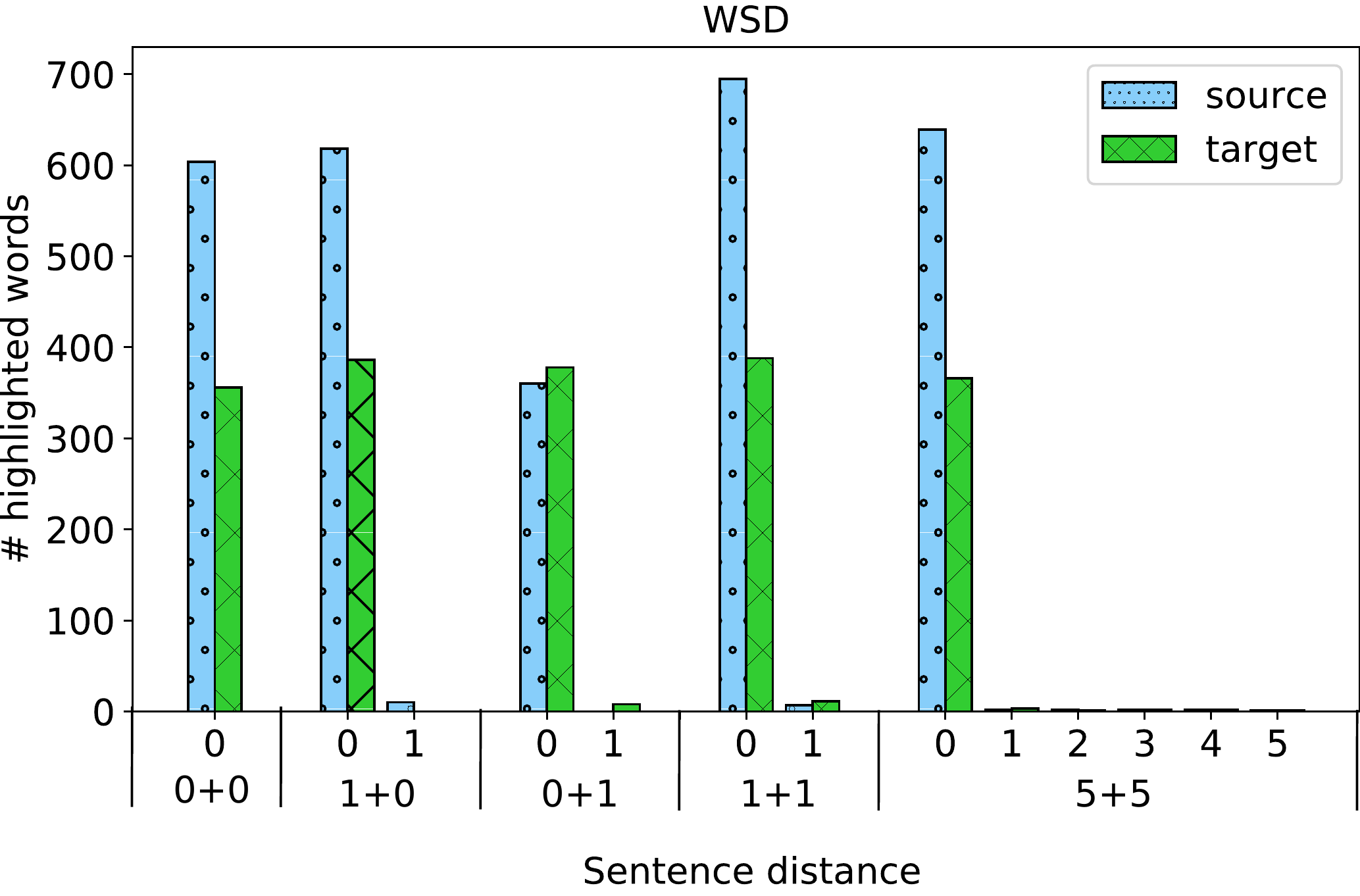}
    \caption{Sentence distance of the highlighted words for each context level for PAR and WSD.}
    \label{fig:dist}
\end{figure}

\begin{table}[t]
\begin{subtable}[h]{\linewidth}
\resizebox{\linewidth}{!}{ 
\begin{tabular}{c|ccc|ccc}
\toprule
& \multicolumn{3}{c}{PAR} & \multicolumn{3}{c}{WSD} \\
POS & Source & Target & Total & Source & Target & Total \\
\midrule
noun & 1550 & 2952 & 4502 & 3340 & 937 & 4277 \\
proper noun & 136 & 4056 & 4192 & 192 & 304 & 496 \\
pronoun & 2676 & 389 & 3065 & 119 & 204 & 323\\
verb & 247 & 859 & 1106 & 406 & 367 & 773 \\
determiner & 499 & 498 & 997 & 800 & 1091 & 1891\\
auxiliary & 310 & 136 & 446 & 78 & 85 & 163 \\
adjective & 105 & 319 & 424 & 291 & 226 & 517 \\
adposition & 65 & 172 & 237 & 283 & 481 & 764 \\
conjunction & 71 & 63 & 134 & 83 & 92 & 175 \\
numeral & 37 & 39 & 76 & 22 & 440 & 462 \\
particle & 37 & 8 & 45 & 61 & 0 & 61\\
\bottomrule
\end{tabular}}
\caption{Part-of-speech}
\label{table:pos}
\end{subtable}
\hfill
\begin{subtable}[h]{\linewidth}
\resizebox{\linewidth}{!}{ 
\begin{tabular}{c|ccc|ccc}
\toprule
& \multicolumn{3}{c}{PAR} & \multicolumn{3}{c}{WSD} \\
DEP & Source & Target & Total & Source & Target & Total \\
\midrule
antecedent & 1297 & 3853 & 5150 & \textendash & \textendash & \textendash \\
determiner & 304 & 1577 & 1881 & 497 & 407 & 904\\
modifier & 168 & 272 & 440 & 258 & 501 & 759 \\
conjunct & 22 & 58 & 80 & 84 & 45 & 129\\
case marking & 7 & 48 & 55 & 4 & 93 & 97\\
\bottomrule
\end{tabular}}
\caption{Dependency relation}
\label{table:dep}
\end{subtable}
\vspace{-1mm}
\caption{Most frequent part-of-speech and dependency relation of highlighted words. 
}
\vspace{-2mm}
\label{table:posdep}
\end{table}

\begin{table}[t]
\resizebox{\linewidth}{!}{ 
\begin{tabular}{p{.55cm}p{10cm}}
\toprule
PAR & \textit{Listen, these big celebrities, they do it different than anybody else?} Jesus, you know if \textbf{they} knew you had hidden cameras in that bedroom... \\
& \textit{Dis-moi, \inline{\hl{ces}}{DET} \inline{\hl{vedettes}}{ante NOUN}, \inline{\hl{elles}}{PRON} le font différemment des autres? } Bon Dieu, tu te rends compte que si \textbf{elles}/\st{ils} savaient que cette chambre cache des caméras... \\
\midrule
WSD & \textit{Right this way.} Your \textbf{charm} is only \inline{\hl{exceeded}}{VERB} by your \inline{\hl{frankness}}{NOUN}. \\
& \textit{Suivez-moi.} Ton \textbf{charme}/\st{portebonheur} n'a d'égal que ta franchise. \\
\bottomrule
\end{tabular}}
\vspace{-1mm}
\caption{Examples of supporting context.}
\vspace{-2mm}
\label{table:posdepex}
\end{table}

\subsection{Analysis of the Highlighted Words}
\label{section:sc-analysis}
Next, we examine the words that were selected as rationales from several angles.

\paragraph{Distance.}
Figure \ref{fig:dist} shows, for each context level, the number of highlighted words at a given distance (in sentences) from the ambiguous word.
For PAR, when no previous sentences are provided, there are as many selected words from the source as the target context. With inter-sentential context, experts selected more supporting context from the target side. One possible reason is that the source and target sentences on their own are equally descriptive to perform PAR, but one may look for the coreference chain of the anaphoric pronoun in the target context to determine its gender, whereas the same coreference chain in the source context would not necessarily contain gender information. Moreover, the antecedent in the target side is more reliable than the source antecedent, since the antecedent can have multiple possible translations with different genders. For WSD, we find that inter-sentential context is seldom highlighted, which reinforces our previous claim that most supporting context for WSD can be found in the current sentences. 

\paragraph{Part-of-Speech and Dependency.}
We use spaCy \cite{spacy2} to predict part-of-speech (POS) tags of selected words and syntactic dependencies between selected words and the ambiguous word. In Table \ref{table:pos}, we find that nominals are the most useful for PAR, which suggests that human translators look for other referents of the ambiguous pronoun to determine its gender. This is reinforced by Table \ref{table:dep}, where the antecedent of the pronoun is selected the most often. 

For WSD, proper nouns and pronouns are not as important as nouns, probably because they do not carry as much semantic load that indicates the sense of the ambiguous word. Determiners, verbs and adpositions are relatively important since they offer clues on the syntactic dependencies of the ambiguous word on other words as well as its role in the sentence, and modifiers provide additional information about the ambiguous word. 

The main difference between PAR and WSD is that for PAR, the key supporting information is \textit{gender}. The source side does not contain explicit information about the gender of the ambiguous pronoun whereas the target side may contain other gendered pronouns and determiners referring to the ambiguous pronoun. For WSD however, the key supporting information is \textit{word sense}. While the source and target sides contain around the same amount of semantic information, humans may prefer to attend to source sentences that express how the ambiguous word is used in the sentence.

\begin{table*}[t]
\resizebox{\linewidth}{!}{ 
\begin{tabular}{c|ccc|ccc|ccc}
\toprule
 & \multicolumn{3}{c}{baseline}  & \multicolumn{3}{c}{attnreg-rand} & \multicolumn{3}{c}{attnreg-pre} \\
Metric & Enc self & Dec cross & Dec self  & Enc self & Dec cross & Dec self & Enc self & Dec cross & Dec self   \\
\midrule 
Dot Product ($\uparrow$, uniform=0.04) &   0.61  & 0.13 & 0.10  &  0.50& 0.28  & 0.20  &  0.53  & 0.63  & 0.31  \\
KL Divergence ($\downarrow$, uniform=3.6) & 3.2 & 3.6 & 3.6  & 3.2  & 3.2 & 3.5 & 3.2 & 3.2 & 3.5   \\
Probes Needed ($\downarrow$, uniform=12.6)  & 8.5 & 11.3  & 13.5 &  7.6  & 9.0 &  5.8 & 6.9 & 6.0  &  10.0  \\
\bottomrule
\end{tabular}}
\vspace{-1mm}
\caption{Alignment between model attention and \textbf{\texttt{SCAT}}.}
\vspace{-3mm}
\label{table:model-human-attn}
\end{table*}

\section{Do Models Pay the Right Attention?}\label{section:model_attention}

Next, we study NMT models and quantify the degree to which the model's attention is aligned with the supporting context from professional translators.

\subsection{Model}

We incorporate the 5 previous source and target sentences as context to the base Transformer \cite{transformer} by prepending the previous sentences to the current sentence, separated by a $\langle$\textsc{brk}$\rangle$ tag, as proposed by \newcite{tiedemann-scherrer-2017-neural}. 

\subsection{Similarity Metrics}
\label{section:metrics}
To calculate similarity between model attention and highlighted context, we first construct a human attention vector $\alpha_{\text{human}}$, where $1$ corresponds to tokens marked by the human annotators, and $0$ otherwise. 
We compare this vector against the model's attention for the ambiguous pronoun for a given layer and head, $\alpha_{\text{model}}$, across three metrics:

\paragraph{Dot Product.} 
The dot product ${\alpha_{\text{human}} \cdot \alpha_{\text{model}}}$ measures the total attention mass the model assigns to words highlighted by humans. 

\paragraph{KL Divergence.} 
We compute the KL divergence between the model attention and the normalized human attention vector $\text{KL}(\alpha_{\text{human-norm}} || \alpha_{\text{model}}(\theta))$, where the normalized distribution $\alpha_{\text{human-norm}}$ is uniform over all tokens selected by humans and a very small constant $\epsilon$ elsewhere such that the sum of values in $\alpha_{\text{human-norm}}$ is equal to $1$.  

\paragraph{Probes Needed.} 
We adapt the ``probes needed'' metric by \citet{faithful-attn} to measure the number of tokens we need to probe, based on the model attention, to find a token highlighted by humans. This corresponds to the ranking of the first highlighted token after sorting all tokens by descending model attention. The intuition is that the more attention the model assigns to supporting context, the fewer probes are needed to find a supporting token.

\subsection{Results}

We compute the similarity between the model attention distribution for the ambiguous pronoun and the supporting context from 1,000 \textbf{\texttt{SCAT}} samples. In Table \ref{table:model-human-attn}, for each attention type we report the best score across layers and attention heads. We also report the alignment score between a uniform distribution and supporting context for comparison. We find that although there is a reasonably high alignment between encoder self attention and \textbf{\texttt{SCAT}}, decoder attentions have very low alignment with \textbf{\texttt{SCAT}}. 




\section{Making Models Pay the Right Attention}
\label{section:attn-reg}
\subsection{Attention Regularization}
We hypothesize that by encouraging models to increase attention on words that humans use to resolve ambiguity, translation quality may improve. We apply \textbf{attention regularization} to guide model attention to increase alignment with the supporting context from \textbf{\texttt{SCAT}}. To do so, we append the translation loss with an \textit{attention regularization loss} between the normalized human attention vector $\alpha_{\text{human-norm}}$ and the model attention vector for the corresponding ambiguous pronoun $\alpha_{\text{model}}$:
$$
\mathcal{R}(\theta) = -\lambda\text{KL}(\alpha_{\text{human-norm}} || \alpha_{\text{model}}(\theta))
$$
where $\lambda$ is a scalar weight parameter for the loss.

During training, we randomly sample batches from \textbf{\texttt{SCAT}} with $p = 0.2$. We train with the standard MT objective on the full dataset, and on examples from \textbf{\texttt{SCAT}}, we additionally compute the attention regularization loss.  

\subsection{Data}

For document translation, we use the English and French data from OpenSubtitles2018 \cite{lison-etal-2018-opensubtitles2018}, which we clean then split into 16M training, 10,036 development, and 9,740 testing samples.
For attention regularization, we retain examples from \textbf{\texttt{SCAT}} where 5+5 context was given to the annotator. We use 11,471 examples for training and 1,000 for testing. 

\subsection{Models}


We first train a \textbf{baseline} model, where the 5 previous source and target sentences serve as context and are incorporated via concatenation. This baseline model is trained without attention regularization. We explore two models with attention regularization: (1) \emph{\textbf{attnreg-rand}}, where we jointly train on the MT objective and regularize attention on a randomly initialized model; (2) \emph{\textbf{attnreg-pre}}, where we first pre-train the model solely on the MT objective, then we jointly train on the MT objective and regularize attention. We describe the full setup in Appendix \ref{appendix:exp}.

\subsection{Evaluation}
As described in Section \ref{sec:background}, we evaluate translation outputs with BLEU and COMET. In addition, to evaluate the direct translation quality of specific phenomena, we translate the 4,015 examples from \newcite{lopes-etal-2020-document} containing ambiguous pronouns that were not used for attention regularization, and we compute the mean word f-measure of translations of the ambiguous pronouns and other words, with respect to reference texts.

We also perform contrastive evaluation on the same subset of \newcite{lopes-etal-2020-document} with a context window of $5$ sentences (\textit{Big-PAR}) and the contrastive test sets by \newcite{bawden-etal-2018-evaluating}, which include 200 examples on anaphoric pronoun translation and 200 examples on lexical consistency/word sense disambiguation. The latter test sets were crafted manually, have a context window of $1$ sentence, and either the previous source or target sentence is necessary to disambiguate.

Context-aware models often suffer from error propagation when using previously decoded output tokens as the target context \cite{li2020does}. Therefore, during inference, we experiment with both using the gold target context (\textbf{Gold}) as well as using previous output tokens (\textbf{Non-Gold}).

\begin{table*}[t]
\resizebox{\linewidth}{!}{ 
\begin{tabular}{c|ccccccccc}
\toprule
 & \multicolumn{2}{c}{\textbf{Gold}}  & \multicolumn{2}{c}{\textbf{Non-Gold}} & \multicolumn{2}{c}{\textbf{F-measure}} & \multicolumn{3}{c}{\textbf{Contrastive Evaluation}}  \\
\textbf{Model} &BLEU & COMET &BLEU & COMET & Pronouns & Other &Big-PAR & PAR & WSD  \\
\midrule
baseline & 33.5  & 41.7 & 29.0 & 37.1 & 0.36 & 0.44 & 90.0 & 60.0 & 55.0\\
attnreg-rand & \bfseries 33.9  & 42.3 &\bfseries 30.2 & 39.3 & \bfseries 0.42 & 0.44 & \bfseries 91.1 & \bfseries 69.0 & 56.0 \\
attnreg-pre & 32.3 & 36.9 & 29.3 & 34.6 & \bfseries 0.48 & 0.44 &  90.6 &  62.5 & 52.5\\

\bottomrule
\end{tabular}}
\vspace{-1mm}
\caption{Overall results. Scores significantly better than baseline with $p < 0.05$ are bolded. COMET scores are multiplied by 100.}
\vspace{-2mm}
\label{table:main_results}
\end{table*}


\subsection{Overall Performance}
Before delving into the main results, we note that we explored regularizing different attention vectors in the model (Appendix \ref{appendix:models}) and obtain the best BLEU and COMET scores for \textit{attnreg-rand} when regularizing the self-attention of the top encoder layer, cross-attention of the top decoder layer and self-attention of the bottom decoder layer. 
For \textit{attnreg-pre}, regularizing self-attention in the top decoder layer gives the best scores.
Thus, we use these as the default regularization methods below.

Moving on to the main results in Table \ref{table:main_results},  we observe that \textit{attnreg-rand} improves on all metrics, which demonstrates that attention regularization is an effective method to improve translation quality. Although \textit{attnreg-pre} does not improve general translation scores significantly, it yields considerable gains in word f-measure on ambiguous pronouns and achieves some improvement over the baseline on contrastive evaluation on Big-PAR and PAR. Attention regularization with supporting context for PAR seems to especially improve models on similar tasks. The disparity between BLEU/COMET scores and targeted evaluations such as word f-measure and contrastive evaluation further suggests that general MT metrics are somewhat insensitive to improvements on specific discourse phenomena. For both models with attention regularization, there are no significant gains in WSD. As discussed in \S\ref{section:sc-analysis}, WSD and PAR require different types of supporting context, so it is natural that regularizing attention using supporting context extracted from only one task does not always lead to improvement on the other.

\subsection{Analysis}

We now investigate how models trained with attention regularization handle context differently compared to the baseline model. 

\paragraph{How does attention regularization influence alignment with human rationales?}
We revisit the similarity metrics from \S\ref{section:metrics} to measure alignment with \textbf{\texttt{SCAT}}. In Table \ref{table:model-human-attn}, the dot product alignment over attention in the decoder increases with attention regularization, suggesting that attention regularization guides different parts of the model to pay attention to useful context. Interestingly, although only the encoder self-attention was explicitly regularized for \textit{attnreg-pre}, the model seems to also have learned better alignment for attention in the decoder. Moreover, \textit{attnreg-pre} generally has better alignment than \textit{attnreg-rand}, suggesting that models respond more to attention regularization once it has been trained to perform translation.


\begin{table}[t]
\resizebox{\linewidth}{!}{ 
\begin{tabular}{ccccccccc}
\toprule
& \textbf{Attention}& \textbf{B} & \textbf{C} & \multicolumn{2}{c}{\textbf{F-meas}} & \multicolumn{3}{c}{\textbf{Contrastive Evaluation}} \\
& & & & Pron.& Other & Big-PAR & PAR & WSD \\
\midrule
AR & Enc self & \bfseries33.8 & \bfseries42.5 & 0.47 & 0.44 &  91.3& 59.5 &  \bfseries 57.0\\
AR & Dec cross & 33.8 & 41.7& 0.47 & 0.43 & 89.8 & 65.0 & 55.5 \\
AR & Dec self & 33.1 & 41.6 & 0.41 & \bfseries0.45 & 88.9 & 66.0 & 53.5\\
AP & Enc self  & 32.3 & 37.4 &0.48 & 0.44 & 91.5 & \bfseries 66.5 & 52.5\\
AP & Dec cross &32.3& 37.4 & 0.47 & 0.43 & \bfseries92.1 & 63.0 &   55.0 \\
AP & Dec self & 32.3 &  36.9 &\bfseries 0.48 & 0.44 & 90.6 & 62.5 & 52.5 \\
\bottomrule
\end{tabular}}
\vspace{-1mm}
\caption{Performance of models with various regularized attention. AR: attnreg-rand, AP: attnreg-pre, B: BLEU, C:COMET}
\label{table:many-attn}
\end{table}

\paragraph{Which attention is the most useful?}
For each of \textit{attnreg-rand} and \textit{attnreg-pre}, we perform attention regularization on either the encoder self-attention, decoder cross-attention or decoder self-attention only. In Table \ref{table:many-attn}, encoder self-attention seems to contribute the most to both translation performance and contrastive evaluation. Although \textit{attnreg-rand} models achieve higher BLEU and COMET scores, \textit{attnreg-pre}  obtain higher scores on metrics targeted to pronoun translation. Attention regularization seems to have limited effect on WSD performance, the scores vary little between attention types.

\begin{table}[t]
\begin{center}
\resizebox{\linewidth}{!}{ 
\begin{tabular}{c|ccc}
\toprule
mask & baseline & attnreg-rand & attnreg-pre \\
\midrule
no mask & 82.5& 84.4 & 86.7 \\
supporting & 75.1 & 69.4 & \underline{55.0}\\
random & 76.0 & 77.0 & 80.4\\
source & 65.5 & 67.3 & 73.0  \\
target & 70.6 & 75.3 & 67.6 \\
all & \underline{65.3} & \underline{67.1} & 68.7  \\
\bottomrule
\end{tabular}}
\end{center}
\vspace{-1mm}
\caption{Contrastive performance with various masks on the context. The lowest score for each model is underlined.}
\vspace{-2mm}
\label{table:mask}
\end{table}

\paragraph{How much do models rely on supporting context?}
We compare model performance on contrastive evaluation on \textbf{\texttt{SCAT}} when it is given full context, and when we mask either the supporting context, random context words with $p = 0.1$, the source context, the target context, or all of the context. 
In Table \ref{table:mask}, we find that \textit{baseline} varies little when the supporting context is masked, which again suggests that context-aware baselines do not use the relevant context, although they do observe a drop in contrastive performance when the source and all context are masked. Models with attention regularization, especially \textit{attnreg-pre} observe a large drop in contrastive performance when supporting context is masked, which indicates that they learned to rely more on supporting context. Furthermore, for \textit{attnreg-pre}, the score after masking supporting context is significantly lower than when masking all context, which may indicate that having irrelevant context can have an adverse effect.
Another interesting finding is that both \textit{baseline} and \textit{attnreg-rand} seem to rely more on the source context than the target context, in contrast to human translators. This result corroborates prior results where models have better alignment with supporting context on attention that attends to the source (encoder self-attention and decoder cross-attention), and regularizing these attention vectors contributes more to translation quality than regularizing the decoder self-attention.


\section{Related Work} 
\label{section:related-work}


\subsection{Context-Aware Machine Translation}
Most current context-aware NMT approaches enhance NMT by including source- and/or target-side surrounding sentences as context to the model. \citet{tiedemann-scherrer-2017-neural} concatenate the previous sentences to the input; \citet{jean2017does, bawden-etal-2018-evaluating, zhang-etal-2018-improving} use an additional encoder to extract contextual features; \citet{wang-etal-2017-exploiting-cross} use a hierarchical RNN to encode the global context from all previous sentences; \citet{maruf-haffari-2018-document, Tu2018LearningTR} use cache-based memories to encode context; \citet{miculicich-etal-2018-document, maruf-etal-2019-selective} use hierarchical attention networks; \citet{chen-etal-2020-modeling-discourse} add document-level discourse structure information to the input. While \citet{maruf-etal-2019-selective, voita-etal-2018-context} also find higher attention mass attributed to relevant tokens in selected examples, our work is the first to guide model attention in context-aware NMT using human supervision and analyze its attention distribution in a quantitative manner. 

However, recent studies suggest that current context-aware NMT models often do not use context meaningfully. \citet{kim-etal-2019-document} claim that improvements by context-aware models are mostly from regularization by reserving parameters for context inputs, and \citet{li-etal-2020-multi-encoder} show that replacing the context in multi-encoder models with random signals leads to similar accuracy as using the actual context. Our work addresses the above disparities by collecting human supporting context to regularize model attention heads during training.

\subsection{Attention Mechanisms}
Though attention is usually learned in an unsupervised manner, recent work supervises attention with word alignments \citep{mi-etal-2016-supervised, liu-etal-2016-neural}, event arguments and trigger words \citep{liu-etal-2017-exploiting, zhao-etal-2018-document}, syntactic dependencies \citep{strubell-etal-2018-linguistically}
or word lexicons \citep{zou-etal-2018-lexicon}. Our work is closely related to a large body of work that supervises attention using human rationales for text classification \citep{barrett-etal-2018-sequence, bao-etal-2018-deriving, faithful-attn, choi-etal-2020-less, pruthi2020evaluating}. Our work, however, is the first to collect human evidence for document translation and use it to regularize the attention of NMT models. 


\section{Implications and Future Work}
In this work, we collected a corpus of supporting context for translating ambiguous words. We examined how baseline context-aware translation models use context, and demonstrated how context annotations can improve context-aware translation accuracy. 
While we obtain promising results for context-aware translation by testing one method for attention regularization, our publicly available \textbf{\texttt{SCAT}} dataset could enable future research on alternative attention regularizers. Moreover, our analyses demonstrate that humans rely on different types of context for PAR and WSD in English-French translation, similar user studies can be conducted to better understand the usage of context in other ambiguous discourse phenomena, such as ellipsis, or other language pairs. We also find that regularizing attention using \textbf{\texttt{SCAT}} for PAR especially improves anaphoric pronoun translation, suggesting that  supervising attention using supporting context from different tasks may help models resolve other types of ambiguities.

One caveat regarding our method for collecting supporting context from humans is the difference between \textit{translation}, translating text from the input, and \textit{disambiguation}, choosing between translation candidates.
During translation, humans might pay more attention to the source sentences to understand the source material, but during disambiguation, we have shown that human translators rely more often on the target sentences. One reason why the model benefits more from increased attention on source may be because the model is trained and evaluated to perform translation, not disambiguation. A future step would be to explore alternative methods for extracting supporting context, such as eye-tracking during translation \cite{obrien2009eye}.



\section{Acknowledgements}
We would like to thank Emma Landry, Guillaume Didier, Wilson Jallet, Baptiste Moreau-Pernet, Pierre Gianferrara, and Duy-Anh Alexandre for helping with a preliminary English-French translation study. We would also like to thank Nikolai Vogler for the original interface for data annotation, and the anonymous reviewers for their helpful feedback. This work was 
supported by the European Research Council (ERC StG DeepSPIN 758969),
by the P2020 programs MAIA and Unbabel4EU (LISBOA-01-0247-FEDER-045909 and LISBOA-01-0247-FEDER-042671), and by the Funda\c{c}\~ao para a Ci\^encia e Tecnologia 
through contract UIDB/50008/2020.

\bibliographystyle{acl_natbib}
\bibliography{anthology,acl2021}
\clearpage
\appendix

\section{\textbf{\texttt{SCAT}}: Supporting Context for Ambiguous Translations }
\label{appendix:scat}




The dataset contains 19,372 annotations in total on 14,000 unique examples of ambiguous anaphoric pronoun translations. In Table \ref{table:data}, for each context level, we report the total number of examples, the overall  answer accuracy, the percentage of \textit{not at all} confident answers, and the number of examples that contain a highlighted word the current/context source/target sentences.

\begin{table}[t]
\resizebox{\linewidth}{!}{ 
\begin{tabular}{c|ccccc}
\toprule
& 0+0 & 1+0 & 0+1 & 1+1 & 5+5 \\
\midrule 
\# Examples & 1,709 & 1,616 & 1,689 & 1,742 & 12,616 \\
Answer accuracy (\%)& 78.4 & 90.7 & 93.31 & 93.7 & 97.2 \\
Not at all confident (\%) & 27.6 & 13.9 & 9.4 & 6.9 & 2.8 \\
Highlighted current source & 953 & 636 & 574 & 533 & 1,169 \\
Highlighted current target & 1,019 & 913 & 941 & 965 & 6,198 \\
Highlighted context source & \textendash & 500 & \textendash & 226 & 711\\
Highlighted context target & \textendash & \textendash & 619 & 709 & 6,011 \\

\bottomrule
\end{tabular}}
\vspace{-1mm}
\caption{Statistics on \textbf{\texttt{SCAT}} for various context levels}
\vspace{-2mm}
\label{table:data}
\end{table}

\begin{table}[t]
\resizebox{\linewidth}{!}{ 
\begin{tabular}{lcl}
\toprule
Source & Target & Class \\
\midrule 
nail & \begin{tabular}{c}clou, ongle \\\textit{metal nail, fingernail}\end{tabular}  & non-synonymous \\
\midrule 
fork & \begin{tabular}{c}diapason, fourche, fourchette \\\textit{tuning fork, pitchfork, kitchen fork}\end{tabular} & non-synonymous \\
\midrule 
mistakes & \begin{tabular}{c}erreurs, fautes \\\textit{errors, faults}\end{tabular} & synonymous\\
\midrule 
heater & \begin{tabular}{c}chauffage, radiateur \\\textit{heating, radiator}\end{tabular} & synonymous \\
\bottomrule
\end{tabular}}
\vspace{-1mm}
\caption{Examples of ambiguous word groups.}
\vspace{-2mm}
\label{table:wsd}
\end{table}

\begin{table*}[t]
\resizebox{\textwidth}{!}{ 
\begin{tabular}{ccccccccccc}
\toprule
\textbf{Model} & \textbf{Attention and Layer}& \multicolumn{2}{c}{\textbf{Gold}}  & \multicolumn{2}{c}{\textbf{Non-Gold}} & \multicolumn{2}{c}{\textbf{F-measure}} & \multicolumn{3}{c}{\textbf{Contrastive Evaluation}} \\
& & BLEU&COMET &BLEU &COMET & Pronouns & Other & Big-PAR & PAR & WSD \\
\midrule
AR & Enc self L1 & 33.8 & 42.5 & 29.9 & 39.0 & 0.47& 0.44 & 91.3 & 59.5 & 57.0 \\
AR & Dec cross L1 & 33.8 & 41.7& 30.5 & 39.3 & 0.47 & 0.43 & 89.8 & 65.0 & 55.5 \\
AR & Dec self L1 & 33.1 & 41.6 & 29.7 & 38.6 & 0.41 & 0.45 & 88.9 & 66.0 & 53.5\\
AR & Enc self L1 + Dec cross L1 + Dec self L6 &  33.9  & 42.3 &30.2 & 39.3 &  0.42 & 0.44 &  91.1 &  69.0 & 56.0 \\
\midrule
AP & Enc self L1  &  32.3 & 37.4 & 28.8 & 34.4  &0.48 & 0.44 & 91.5 &  66.5 & 52.5\\
AP & Dec cross L1 & 32.3& 37.4 & 28.7 & 34.2 & 0.47 & 0.43 & 92.1 & 63.0 &   55.0 \\
AP & Dec self L1 & 32.3 &  36.9 & 29.3 & 34.6 & 0.48 & 0.44 & 90.6 & 62.5 & 52.5 \\
AP & Enc self L1 + Dec cross L1 + Dec self L6 & 32.3 & 36.9 & 29.3 & 34.6 & 48.0 & 44.1 & 90.6 & 62.5 & 52.5 \\
\bottomrule
\end{tabular}}
\vspace{-1mm}
\caption{Results of all models with regularized attention. AR: attnreg-rand, AP: attnreg-pre}
\label{table:all-attn}
\end{table*}

\section{Generating Data for Word Sense Disambiguation}
\label{appendix:wsd}

To automatically generate contrastive examples of WSD, we identify English words that have multiple French translations. To do so, we first extract word alignments from OpenSubtitles2018 using AWESOME-align \citep{dou2021word} and obtain:
\[
A_m=\{ \langle x_i, y_j\rangle: x_i \in \xv_m, y_j \in \yv_m \},
\] where for each word pair $\langle x_i, y_j\rangle$, $x_i$ and $y_j$ are semantically similar to each other in context. 

For each pair $\langle x_i, y_j \rangle \in A_m$,  we compute the number of times the lemmatized source word type  $(v_x\!=\!\text{lemma}(x_i))$ along with its POS tag $(t_x\!=\!\text{tag}(x_i))$ is aligned to the lemmatized target word type $(v_y\!=\!\text{lemma}(y_j))$: $c(v_x,t_x,v_y)$. 
Then, we extract tuples of source types with its POS tags  $\langle v_x, t_x\rangle$ that have at least two target words that have been aligned at least 50 times ($|\{v_y | c(v_x,t_x,v_y) \ge 50\}| \ge 2$). 
Finally, we filter out the source tuples which have an entropy $H(v_x, t_x)$ less than a pre-selected threshold $z$. This entropy is computed using the  conditional probability of a target translation given the source word type and its POS tag as follows:
    \[ p:= p(v_y|v_x, t_x) = \frac{c(v_x,t_x,v_y)}{c(v_x,t_x)} \]
\[ H(v_x, t_x) = \sum_{v_y \in \text{trans}(v_x, t_x)}  -p\log_e p \] where  $\text{trans}(v_x, t_x)$ is the set of target translations for the source tuple $\langle v_x, t_x \rangle$ and $p(v_y|v_x, t_x)$ is the conditional probability of a given target translation $v_y$ for the source word type $v_x$ and its POS tag $t_x$

Out of the $394$ extracted word groups, we manually validate and retain 201 groups and then classify them into 64 synonymous and 137 non-synonymous word groups (Table \ref{table:wsd}). We create contrastive translations by extracting sentence pairs containing an ambiguous word pair, and replacing the translation of the polysemous English word by a different French word in the same group. For word groups with synonymous French words, we only retain examples where the French word appears within the previous $5$ sentences to enforce lexical consistency, as otherwise the different French words may be interchangeable. 

\section{Experimental Setup}
\label{appendix:exp}
\subsection{Data preprocessing}
We use the English and French data from the publicly available OpenSubtitles2018 dataset \cite{lison-etal-2018-opensubtitles2018}. We first clean the data by selecting sentence pairs with a relative time overlap between source and target language subtitle frames of at least 0.9 to reduce noise. 
Each data is then encoded with byte-pair encoding \cite{sennrich-etal-2016-neural} using SentencePiece \cite{kudo-richardson-2018-sentencepiece}, with source and target vocabularies of 32k tokens.

\subsection{Training configuration}
We follow the Transformer base \cite{transformer} configuration in all our experiments, with $N = 6$ encoder and decoder layers, $h = 8$ attention heads, hidden size $d_\text{model} = 512$ and feedforward size $d_\text{ff} = 2048$. We use the learning rate schedule and regularization described in \citet{transformer}. We train using the Adam optimizer \cite{DBLP:journals/corr/KingmaB14} with $\beta_1 = 0.9, \beta_2 = 0.98$.

\subsection{Attention regularization setups}
\label{appendix:models}
For both \textit{attnreg-rand} and \textit{attnreg-pre}, we experiment performing regularization on different model attentions at different layers. For the attention regularization loss, In all experiments, we compute the attention regularization loss on the first attention head with $\lambda = 10$, and we divide the loss by the length of the input. We give the results for all setups in Table \ref{table:all-attn}.

\subsection{Evaluation}
To calculate BLEU scores we use SacreBLEU BLEU+case.mixed+numrefs.1+smooth.exp+tok.13 a+version.1.4.14 \cite{post-2018-call} and for COMET we use \textit{wmt-large-da-estimator-1719} \footnote{\url{https://github.com/Unbabel/COMET}}. We test for statistical significance with $p < 0.05$ using bootstrap sampling on a single run \cite{koehn-2004-statistical}.

\end{document}